\documentclass{article}
\usepackage{ijcai17}
\usepackage{times}
\usepackage{helvet}
\usepackage{tikz}
\usepackage{enumerate}
\usepackage{courier}
\usepackage{hyperref}
\usepackage{url}
\usepackage{amsmath,bm}
\usepackage{graphicx}
\usepackage{diagbox}
\usepackage{gensymb}
\usepackage{makecell}

\usepackage{amsmath}
\usepackage{amssymb}
\DeclareMathOperator*{\argmin}{argmin}
\usepackage{algorithm}
\usepackage{subfigure}
\usepackage{multirow}
\usepackage[multiple]{footmisc}
\usepackage[small]{caption}
\usepackage{mathtools}

\DeclarePairedDelimiter\floor{\lfloor}{\rfloor}

\newcommand{\Rmnum}[1]{\expandafter\@slowromancap\romannumeral #1@}

\setboolean{@twoside}{false}

\title{A Survey on Methods and Theories of Quantized Neural Networks}

\author{\textnormal{Yunhui Guo} \\ 
	University of California, San Diego \\
	Computer Science and Engineering Department \\
	yug185@eng.ucsd.edu
}
\hypersetup{draft}
\begin{document}
\maketitle

\begin{abstract}
Deep neural networks are the state-of-the-art methods for many real-world tasks, such as computer vision, natural language processing and speech recognition. For all its popularity, deep neural networks are also criticized for consuming a significant amount of memory and draining battery life of devices during training and inference. This makes it hard to deploy these models on mobile or embedded devices which have tight resource constraints. Quantization is recognized as one of the most effective approaches to satisfy the extreme memory requirements that deep neural network models demand. Instead of adopting 32-bit floating point format to represent weights, quantized representations store weights using more compact formats such as integers or even binary numbers. Despite a possible degradation in predictive performance, quantization provides a potential solution to greatly reduce the model size and the energy consumption. In this survey, we give a thorough review of different aspects of quantized neural networks. Current challenges and trends of quantized neural networks are also discussed. 

\end{abstract}

\section{Introduction}
Since the success on the ImageNet dataset \cite{krizhevsky2012imagenet}, deep neural networks has drawn a huge amount of attention from academia, industry and media. Subsequent works show that deep neural network models can achieve the state-of-the-art results on many real-world tasks, such as computer vision \cite{he2016deep}, natural language processing \cite{young2017recent} and speech recognition \cite{hinton2012deep}. One constraint that hinders the wide use of deep neural network models is that it consumes a huge amount of memory to store the models. For example, AlexNet \cite{krizhevsky2012imagenet} requires 200MB memory, VGG-Net \cite{simonyan2014very} requires 500MB memory and ResNet-101 \cite{he2016deep} requires 200MB memory. For mobile or embedded devices that do not have enough memory space, it is hard to deploy these models into production stack. Quantization is a potential solution to this problem. Quantized neural networks represent weights, activations or even gradients with a small numbers of bits, such as 8 bits or even 1 bit. In this way, we can effectively shrink the model size and accelerate both the training and the inference procedures. 

Quantizing neural networks dates back to the 1990s \cite{fiesler1990weight,balzer1991weight,tang1993multilayer,marchesi1993fast}. In the early days, the main reason to quantize these models is to make it easier for digital hardware implementation. Recently, the research of quantizing neural networks has revived due to the success of deep neural networks and their huge sizes. A slew of new quantization methods and methodologies have been proposed. These efforts have enabled the quantized neural networks to have the same accuracy level as their full-precision counterparts. In this paper, we give a thorough survey on methods and approaches of quantized neural networks. We also discuss the challenges of quantizing neural networks and address future trends.

The rest of the paper is organized as follows: Section 2 gives background on neural networks and specially on quantized neural networks. Section 3 introduces some common quantization methods. In section 4, we discuss the quantization of different network components. In section 5, we compare two types of quantization methodologies. Section 6 gives some case studies. Section 7 discusses about why quantized neural networks work well in practice. In section 8 we discuss possible future directions of quantized neural networks.

\begin{table*}[!h]
	\caption{Specifications of four CNN architectures}
	\label{table:exp}
	\begin{center}
		\begin{tabular}{|c|c|c|c|c|}
			\hline
			& \# of parameters & Layers  & flops & Top-1 error rate on ImageNet  \\ \hline
			AlexNet & 60M  & 8 & 725M  & 43.45\%  \\ \hline
			VGGNet-16 (with batch normalization)  &  138M & 16 &   15484M & 26.63\%  \\ \hline
			GoogleNet & 6.9M  & 22 &  1566M &  31.30\%  \\ \hline
			ResNet-152 &  60.2 M & 152 &  11300M & 22.16\%  \\ \hline
			
		\end{tabular}
	\end{center}
\end{table*}
\section{Background}
\subsection{Neural Networks}

\subsubsection{Feed-forward Neural Networks}
\begin{figure}[h]
	\centering	
	
	\includegraphics[width=1.8in,height=2in]{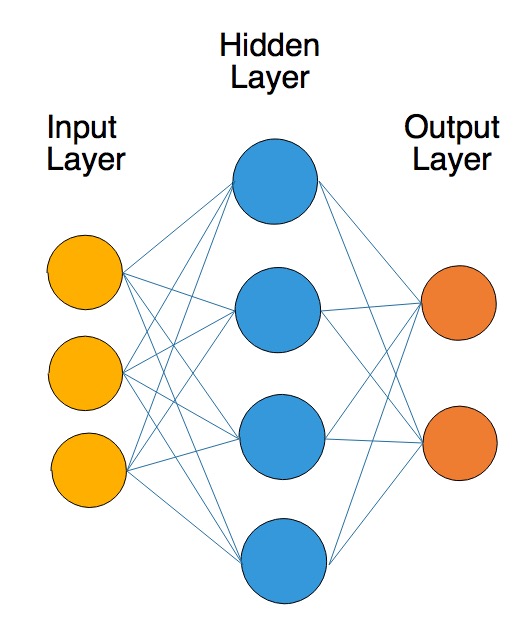} 
	
	\caption{A three-layer feedforward neural network.}
	\label{figure: 1}
\end{figure}

A feed-forward neural network is the artificial neural network without loops. It is the simplest and oldest type of artificial neural network. A feed-forward neural network consists of three parts: input layer, hidden layers and output layer. An example of a feed-forward neural network is shown in Fig \ref{figure: 1}. Usually there are multiple hidden layers. Each hidden layer transforms its input to some representations that the next layer can compute. We use $w^l_{ij}$ to denote the weight from node $i$ in layer $l-1$ to the node $j$ in layer $l$. In a fully connected feed-forward neural network, given an input vector $x^{l-1} \in \mathbb{R}^m$ in layer $l-1$, the output of node $j$ in next layer can be computed as $z^l_j = g(\sum_{i=0}^{m}w^l_{ij}x^{l-1}_i)$, where $g(x)$ is an activation function.

If there are $m$ neurons in layer $l-1$ and $n$ neurons in layer $l$, with fully-connected layer the weights between the two layers are represented as an $m \times n$ matrix. This matrix consumes a large amount of memory when $m$ or $n$ is too large. For example, if a gray-scale input image of the neural network is of size $256 \times 256$ and the first hidden layer has $512$ neurons. Storing the weight matrix with floating-point numbers requires 128M memory. In practice, the images are much larger than this example so a fully-connected layer cannot scale well.

\subsubsection{Convolutional Neural Networks}
\begin{figure}[h]
	\centering	
	\includegraphics[width=1.8in,height=2in]{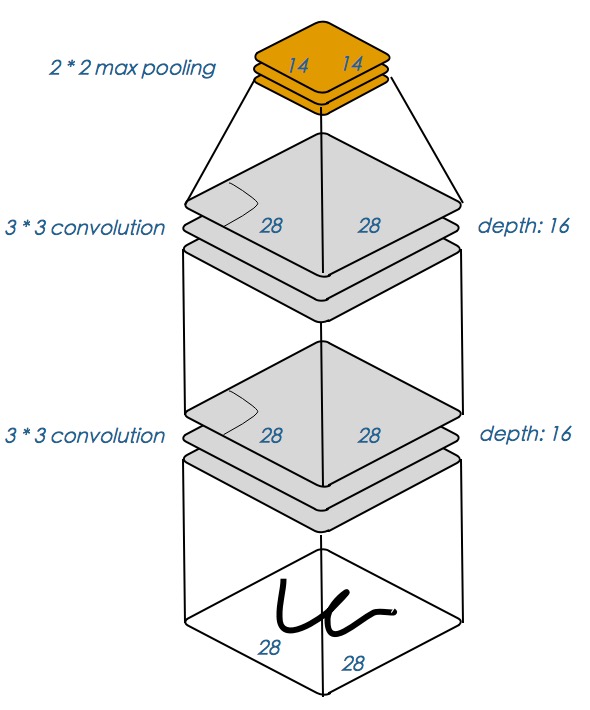} 
	
	\caption{An illustration of convolutional neural networks.}
	\label{figure: 2}
\end{figure}

Convolutional neural network (CNN) is a type of artificial neural network that has been successfully applied in many areas, especially in visual imagery \cite{lecun1998gradient}. A convolutional neural network consists of three building blocks: convolutional layer, pooling layer and fully-connected layer. A simple convolutional neural network is shown in Fig \ref{figure: 2}.

Convolutional layer is a major building block of CNNs. It is used to extract features from images. In each convolutional layer we have a set of filters. During the forward pass, we slide each filter across the image and compute dot products between the filter and the local receptive field. The output of the convolutional layer is called activation map that gives the response of each filter. Given an image $I$ and a $m \times n$ filter $F$, an element $a_{i,j}$ in the activation map can be computed as, 
\begin{equation}
	a_{i,j} = \sum_{a = 1}^{m}\sum_{b = 1}^{n} I_{i+a-1,j+b-1} \times F_{a,b}
\end{equation}

The convolution operation is computationally very expensive. For example, the total time complexity of all convolutional layers can be expressed as $O(\sum_{l=1}^d n_{l-1} \cdot s_{l}^2 \cdot n_{l} \cdot m_l^2)$ \cite{he2015convolutional}. Here $l$ is the index of a convolutional layer and $d$ is the number of convolutional layers. $n_l$
is the number of filters in the $l$-th layer. $n_{l-1}$ is the number of input channels of the $l$-th layer. $s_l$ is the spatial size of the filter. $m_l$ is the spatial size of the activation map. The computational cost of the convolutional layer motivates us to use low bit-width filters and inputs. With low bit-width filters and inputs, the dot product can be efficiently implemented by bitwise operations which can greatly accelerate the computation.

The success of Alexnet \cite{krizhevsky2012imagenet}
at ILSVRC 2012 spawned a lot of novel CNN architectures. In this paper, we focus on the following four CNN architectures,
\begin{itemize}
	\item AlexNet \cite{krizhevsky2012imagenet}
	\item VGGNet \cite{simonyan2014very}
	\item GoogleNet \cite{szegedy2015going}
	\item ResNet \cite{he2016deep}
\end{itemize}

The performance of these models is impressive, however their huge size hinders them from being widely used. This motivates researchers to develop quantization methods to further reduce the model size. The four architectures are widely used as baselines to compare the effectiveness of different quantization approaches. The specifications of these models are given in Table 1. More details can be found in the corresponding papers.

\begin{figure}[h]
	\centering	
	
	\includegraphics[height=1.4in]{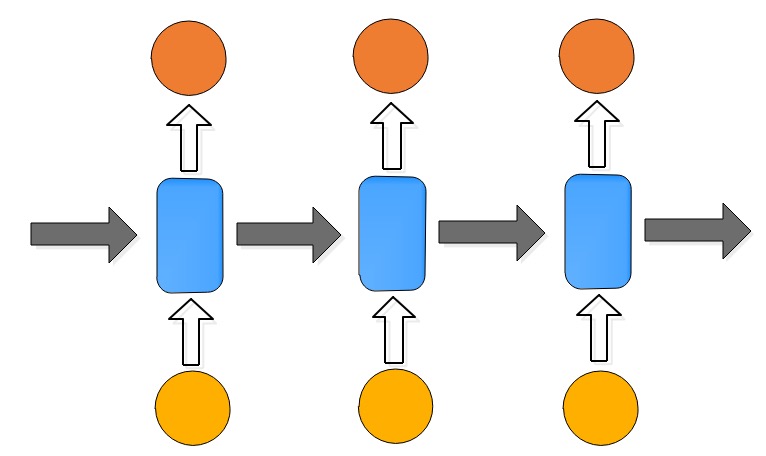} 
	
	\caption{A recurrent neural network.}
	\label{figure: rnn}
\end{figure}

\subsubsection{Recurrent Neural Networks (RNNs) and LSTM}
Recurrent neural networks (RNNs) and Long Short-Term Memory (LSTM) \cite{hochreiter1997long} are used to model the dynamics of sequences. Different from feed-forward neural networks and convolutional neural networks (CNNs), RNNs and LSTM may include loops that are used to consider the previous computations. An example of RNN is shown in Fig \ref{figure: rnn}. The motivation to quantize RNNs and LSTM is not fundamentally different from quantizing feed-forward neural networks and CNNs. In order to achieve satisfactory performances, we need millions of parameters to model complex sequential relations \cite{amodei2016deep} which makes it infeasible to deploy these models into embedded or mobile devices. 

\subsection{Quantized Neural Networks}
The research of quantized neural networks has attracted a lot of attention from the deep learning community \cite{courbariaux2015binaryconnect,rastegari2016xnor,zhou2017incremental}. The goal of quantization is to compact the models without performance degradation. Achieving this goal calls for joint solutions from machine learning, optimization computer architecture, and hardware design, etc. With quantized neural networks, we can use bitwise operations rather than floating-point operations to perform the forward and back-propagation. A simple example is that for two binary vectors, their dot product can be computed as follows,
\begin{equation}
	\textbf{a} \cdot \textbf{b} = \textnormal{bitcount}(\textbf{a}\ \textnormal{and}\ \textbf{b})
\end{equation}

\noindent where bitcount() is a function that counts the number of 1s in a binary vector. We can also save energy with quantized neural networks. The computational operations in quantized neural networks are typically bitwise operations which are carried out by the arithmetic logic unit (ALU). An ALU consumes much less energy than a floating-point unit (FPU). For mobile applications where power consumption is critical, a quantized neural network is preferable over its full precision counterpart.

A lot of techniques have been proposed recently to quantize neural networks. Broadly speaking, these techniques can be classified into two types: deterministic quantization and stochastic quantization. In deterministic quantization, there is an one-to-one mapping between the quantized value and the real value. While in stochastic quantization, the weights, activations or gradients are discretely distributed. The quantized value is sampled from the discrete distributions.

There are three components that can be quantized in a neural network: weights, activations and gradients. The motivation and methods to quantize these components are slightly different. With quantized weights and activations we get smaller model size. In a distributed training environment, we can save communication cost with quantized gradients. Generally, it is more difficult to quantize the gradients than quantizing weights and activations since high-precision gradients are needed to make the optimization algorithm converge.

We use quantization codebook to denote the set of discrete values used to represent the real values. From the perspective of quantization codebook, the works on quantized neural networks can be roughly classified into two categories: fixed codebook quantization and adaptive codebook quantization. In fixed codebook quantization, the weights are quantized into some predefined codebook while in adaptive codebook quantization the codebook is learned from the data. Some commonly used codebooks are $\{-1,1\}$, $\{-1,0,1\}$ or power-of-two numbers, which provides binary network, ternary network and power-of-two network separately.

The recent quantized neural networks have achieved accuracy similar to their full-precision counterparts. For example, a binary network \cite{courbariaux2015binaryconnect} can obtain 98.8\% accuracy on the MNIST dataset. For large datasets such as ImageNet, a ternary network \cite{zhu2016trained} can obtain comparable performance to the full-precision network. However, there are still several challenges need to be addressed. Training quantized networks needs more tuning and the working mechanism of quantized networks are not well understood.  Exploring new quantization methods and developing theories for quantized neural networks are important.

\section{Quantization Techniques}
\subsection{ Deterministic quantization}
\vspace{0.1cm} 
\subsubsection{Rounding}  
Rounding is possibly the simplest way to quantize real values. In \cite{courbariaux2015binaryconnect}, the authors proposed the following deterministic rounding function,
\begin{equation}
x^b = \textnormal{Sign}(x) = \begin{cases}
+1\quad x \ge 0,\\
-1\quad \textnormal{otherwise}\\
\end{cases}
\end{equation}
where $x^b$ is the binarized variable and $x$ is the real-valued variable. This function can produce binarized weights, activations or gradients. During forward propagation, the real-value weights are quantized via the Sign(x) function and the quantized weights are used to generate the outputs. However, during back-propagation we cannot back-propagate the errors through the Sign(x) function since the gradients are zero almost  everywhere. The usual strategy is to use ``straight-through estimator" (STE) \cite{STE} which is a heuristic way to estimate the gradient of a stochastic neuron. Assume $E$ is the loss function, with STE the forward and backward computations of above rounding function are as follows,

\begin{equation}
\begin{split}
& \textbf{Forward:}\quad x^b = \textnormal{Sign}(x) \\
& \textbf{Backward:}\quad \frac{\partial E}{\partial x}  =  \frac{\partial E}{\partial x^b}\textnormal{I}_{|x| \le 1}	
\end{split}
\end{equation}

where $\textnormal{I}_{|x| \le 1}$ is an indicator function defined as,
\begin{equation}
\textnormal{I}_{|x| \le 1} = \begin{cases}
1\quad |x| \le 1,\\
0\quad \textnormal{otherwise}\\
\end{cases}
\end{equation}

In order to round a floating-point number to the nearest fixed-point representation, in \cite{gupta2015deep} the authors proposed the following rounding scheme,

\begin{equation}
\textnormal{Round}(x, [\textnormal{IL},\textnormal{FL}]) =  \begin{cases}
\floor{x} \hskip2.6em \textnormal{if}\ \floor{x} \le x \le \floor{x} + \frac{\epsilon}{2} ,\\
\floor{x}+\epsilon  \hskip1em \textnormal{if}\ \floor{x}+\frac{\epsilon}{2} < x \le \floor{x} + \epsilon \\
\end{cases}
\end{equation} 

In a fixed-point representation, IL represents the number of integer bits and FL represents the number of fractional bits. $\epsilon$ is the smallest positive number that can be represented in this fixed-point format. $\floor{x}$ is defined as the largest integer multiple of $\epsilon$. For values that are beyond the range of this fixed-point format, the authors normalized them to either the lower or the upper bound of the fixed-point representation. \cite{rastegari2016xnor} extended Equation (4) as follows,

\begin{equation}
\begin{split}
& \textbf{Forward:}\quad\hskip0.8em x^b = \textnormal{Sign}(x) \times \textnormal{E}_F(|x|) \\
& \textbf{Backward:}\quad \frac{\partial E}{\partial x}  =  \frac{\partial E}{\partial x^b}	
\end{split}
\end{equation}
where $\textnormal{E}_F(|x|)$ is the mean of absolute weight values of each output channel. In \cite{zhou2016dorefa}, instead of doing a channel-wise scaling, the authors replaced $\textnormal{E}_F(|x|)$ with a constant scalar for all the filters.

More recently, \cite{polino2018model} proposed a general rounding function,
\begin{equation}
	Q(x) = sc^{-1}(\hat{Q}(sc(x))),
\end{equation}

where $sc(x)$ is a scaling function which maps the values from arbitrary range to the values in [0, 1]. $\hat{Q}(x)$ is the actual quantization function. Given a quantization level parameter $s$, the uniform quantization function with $s+1$ levels can be defined as,
\begin{equation}
\hat{Q}(x, s) = \frac{\floor{xs}}{s} + \frac{\xi}{s} 
\end{equation}
where,

\[
\xi = \begin{cases}
1 \quad xs - \floor{xs} > \frac{1}{2},\\
0 \quad \textnormal{otherwise}\\
\end{cases}
\]

The intuition of this quantization function is that $x$ will be assigned to the nearest quantization point of $s-1$ equally spaced points between 0 and 1. This a generalized version of the Sign(x) function and can be used to quantized the real values into multi-levels. In \cite{wu2018training}, the authors proposed a heuristic rounding function to quantize a real value to a k-bit integer,
\begin{equation}
Q(x,k) = Clip\{\sigma(k)\cdot round[\frac{x}{\sigma(k)}], -1+\sigma(k), 1-\sigma(k)\}
\end{equation}

The idea is to quantize real values with uniform distance $\sigma(k)$, where $\sigma(k) = 2^{1-k}$. $Clip$ restricts the quantized values in the range of $[-1+\sigma(k), 1-\sigma(k)]$ and $round$ replaces the continuous values with their nearest discrete points. 
\\

\noindent {\small \textbf{Challenges:}}  Use a rounding function is an easy way to convert real values into quantized values. However, the network performance may drop dramatically after each rounding operation. It is necessary to keep the real values as reference during training which increases the memory overhead. Meanwhile, since the parameter space is much smaller if we use discrete values, it is harder for the training process to converge. Finally, rounding operation cannot exploit the structural information of the weights in the network.

\subsubsection{Vector Quantization}
To the best of our knowledge, \cite{gong2014compressing} is the first paper to systematically consider using vector quantization to quantize and compress neural networks. The basic idea of vector quantization is to cluster the weights into groups and use the centroid of each group to replace the actual weights during inference. 

For a weight matrix $W \in R^{m \times n}$, we can perform k-means clustering to do vector quantization,
\begin{equation}
	\textnormal{min} \sum_{i}^{m}\sum_{j}^n\sum_{k}^l \lVert w_{ij} - c_k \rVert_2^2
\end{equation}
where $c_k$ is the centroid. After clustering, each weight is assigned to a cluster index. Although this is a simple approach, the authors showed that on the ImageNet dataset \cite{imagenet_cvpr09}, this method can achieve 16 $\sim$ 24 times compression of the network with only 1\% loss of
classification accuracy using the state-of-the-art CNNs. In \cite{han2015deep}, the authors adopted a similar approach to \cite{gong2014compressing} except that after clustering, they retrained the network to fine-tune the quantized centroids.

While simple and effective, \cite{choi2016towards} pointed out the above quantization method have two drawbacks. The first one is that we cannot control the loss of accuracy caused by the k-means clustering. The second is that k-means clustering does not impose any compression ratio constraint. To solve the problems, the authors proposed a Hessian-weighted k-means clustering approach. The basic idea is to use the Hessian-weighted distortion to measure the performance degradation that will be caused by the quantization of weights. In this way, those weights have an large impact on the performance of the network are prevented from deviating from their original values too much.

There are many extensions to vector quantization. For example, product quantization \cite{gong2014compressing} is a way that partitions the weight matrix into many disjoint sub-matrices and performs quantization in each sub-matrix. In \cite{wu2016quantized}, the authors adopted product quantization with error correction to quantize network parameters to enable fast training and testing. Residual quantization \cite{gong2014compressing} quantizes the vectors into $k$ clusters and then recursively quantize the residuals. In \cite{park2017weighted}, the authors adopted a way which is similar to vector quantization. They used an idea based on weight entropy \cite{guiacsu1971weighted} to group weights into $N$ clusters. There are more clusters for important ranges of weights. In this way, they can achieve automatic and flexible multi-bit quantization.
\\

\noindent {\small \textbf{Challenges:}} Due to the number of weights in the network, the computation of the k-means clustering is expensive. Compared with rounding methods, it is hard to use vector quantization to achieve binary weights. Vector quantization is typically used for quantizing pre-trained models. Hence, if the task is to train a quantized network from scratch, it is preferable to use a carefully designed rounding functions. Vector quantization ignores the local information of the network.

\subsubsection{Quantization as Optimization}
Recently, a number of works have considered formulating the quantization problem as an optimization problem. In XNOR-net \cite{rastegari2016xnor}, in order to find the best binary approximation to the real-value filters, the authors solved the following optimization problem,
\begin{equation}
	J(B, \alpha) = \lVert W - \alpha B  \rVert^2
\end{equation}

where $W$ is a real-value filter, $B$ is the binary filter and $\alpha$ is a positive scaling factor. The optimal $B$ and $\alpha$ are given as follows,
 
\begin{equation}
B^* = \textnormal{Sign}(W), \quad \alpha^* = \frac{1}{n}\lVert W  \rVert_{l_1}
\end{equation}
where $n$ is the number of the elements in the filter. Interestingly, this gives a result similar to Equation (3) except that in this case there is an additional scaling factor. 

In \cite{li2016ternary}, the authors relaxed the binary constraint to ternary values and solved the following optimization problem,
\begin{equation}
	\alpha^*, W^{t^*} = \begin{cases}
     \argmin_{\alpha, W^t} J(\alpha, W^t) = \lVert W - \alpha W^t  \rVert^2_2\\
	\textnormal{s.t.} \quad \alpha \ge 0, W^t_{i} \in {-1, 0, 1}, i = 1,2,...,n. \\
	\end{cases}
\end{equation}
In this way, higher accuracy can be achieved compared to XNOR-net \cite{rastegari2016xnor}. Instead of fixing the codebook in ternarization, in \cite{zhu2016trained} the authors used a trained quantization method to learn the ternary values which gives the network more flexibility. In \cite{mellempudi2017ternary}, the authors introduced multiple scaling factors into ternary network to account for the unsymmetry between positive and negative weights. In \cite{wang2017fixed}, the authors proposed the following semi-discrete decomposition for a weight matrix $W \in R^{m \times n}$:

\begin{equation}
\textnormal{min}_{X,D,Y}{\lVert W - XDY^T \rVert^2_F}
\end{equation}

where $X \in \{-1, 0, +1\}^{m \times k}$, $Y \in \{-1, 0, +1\}^{n \times k}$ and $D \in R^{k \times k_+}$ is a nonnegative diagonal matrix. By choosing different $k$, we can make a trade-off between compression ratio and performance loss.

A different type of approach is to minimize the loss function directly with respect to the quantized weights. In \cite{hou2016lossaware}, the authors considered the effect of binarization on the loss function during training and proposed a loss-aware binarization algorithm. They formulated the binarization problem as the following optimization problem:
\begin{equation}
	 \begin{cases}
	\textnormal{min}_{\hat{\textbf{w}}} E(\hat{\textbf{w}}) \\
	\textnormal{s.t.} \quad \hat{\textbf{w}}_l = \alpha_l \textbf{b}_l, \alpha_l > 0, \textbf{b}_l \in \{\pm 1\}^{n_l}, l = 1,...,L, \\
	\end{cases}
\end{equation}
where $E$ is the loss function and $n_l$ is the number of the parameters in layer $l$ and $L$ is the number of layers. The problem then is rewritten to a formulation that can be solved by proximal Newton method.

In \cite{carreira2017model}, the authors formulated the quantization problem as following non-convex optimization problem,

\begin{equation}
	\textnormal{min}_{\hat{\textbf{w}}, \textbf{w}} E(\hat{\textbf{w}}) \quad \textnormal{s.t.} \quad \hat{\textbf{w}} = \Delta(\textbf{w})
\end{equation}
where $\textbf{w}$ is the real-valued weights and $\hat{\textbf{w}}$ is the quantized weights. $E(\hat{\textbf{w}})$ is the loss function of the quantized network and $\Delta$ is a quantization function that converts real-valued weights to discrete values. As compared to \cite{hou2016lossaware}, this is a more general setting since it allows us to use different quantization functions. Then they used the ``learning-compression" algorithm to train the network. In \cite{leng2017extremely}, the authors formulated the quantization problem as a discretely constraint non-convex optimization problem and used the idea of Alternating Direction Method of Multipliers (ADMM) to decouple the continuous variables from the discrete constraints. The optimization problem is formally defined as,

\begin{equation}
 \begin{cases}
\textnormal{min}_{\textbf{w}}E(\textbf{w}) \\
\textnormal{s.t.} \quad \textbf{w}_{ij} \in {-2^N, ...,-2^1,-2^0,0,+2^0,+2^1,...,+2^N} \\
\end{cases}
\end{equation}

The authors further introduced a scaling factor to each layer to expand the constraint space. Then the problem is converted into a form that can be solved by ADMM. More recently, \cite{hou2018lossaware} extended \cite{hou2016lossaware} to loss-aware ternarization and $m$-bit quantization. The authors also used proximal Newton algorithm and obtained a closed-form update for the optimization problem.
 
In \cite{zhou2017adaptive}, the authors considered the problem of finding the optimal quantized representation for each layer. Later in the work of \cite{khoram2018adaptive} the authors proposed an adaptive quantization method which incorporates the loss function into an optimization problem to consider the importance of different connections. The optimization problems is as follows,
\begin{equation}
\begin{cases}
\textnormal{min}_{\textbf{w}}N_Q(\textbf{w}) = \sum_{i=1}^n N_q(w_i) \\
\textnormal{s.t.} \quad E(\textbf{w}) \le \bar{E}  \\
\end{cases}
\end{equation}

where $N_q(w_i)$ is the minimum number of bits used to represent $w_i$ and $N_Q(\textbf{w})$ is the total number of bits in the network. $\bar{E}$ is used to bound the accuracy loss. The intuition is to use more precise representation for important weights while allocating few bits for unimportant weights. Different from above works, the work in \cite{deng2017} proposed gated XNOR networks which use discrete state transition method to optimize the weights in discrete space. 
\\

\noindent {\small \textbf{Challenges:}}
The convergence of the proposed optimization algorithms relies on weak assumptions which may not hold for deep neural networks. This makes the theoretical analysis of these algorithms not very convincing. Some of the methods need second-order information for updating the weights which leads to high computational complexity. From a practical perspective, it calls for more efforts to implement the proposed optimization algorithms which hinders their widespread use.

\subsection{Stochastic Quantization}

\subsubsection{Random Rounding}
In random rounding, the real value has no one-to-one mapping to the quantized value. Typically, the quantized value is sampled from a discrete distribution which is parameterized by the real values. For example, in \cite{courbariaux2015binaryconnect} the authors proposed the following random rounding function,
\[
x^b = \begin{cases}
+1\quad \textnormal{with probability}\ p = \sigma(x),\\
-1\quad \textnormal{with probability}\ 1 - p \\
\end{cases}
\]

where $\sigma$ is the "hard sigmoid" function:
\begin{equation*}
\sigma(x) = \textnormal{clip}(\frac{x+1}{2}, 0, 1) = \max(0, \min(1, \frac{x+1}{2}))
\end{equation*}
The intuition is that if $x$ is a positive value, we will have a high probability to quantize it to +1, otherwise to -1. This gives us a more flexible quantization scheme. In \cite{muller2015rounding} the authors used the idea in integer programming. The proposed random rounding function maps each real value probabilistically to  either the nearest discrete point or to the second nearest discrete point based on  the distance to the corresponding point. In \cite{lin2015neural}, the authors extended binary random rounding to the ternary case. They first split the interval $[-1,1]$ into two sub-intervals: $[-1,0]$ and $(0,1]$. If the real-valued weight $w$ is in $[-1,0]$, then the weight is quantized as follows,
\begin{equation}
	P(w^b_{ij} = -1) = w_{ij};\quad  P(w^b_{ij} = 0) = 1- w_{ij}  
\end{equation} 
The case for $w$ $\in$ $(0,1]$ is similar. In \cite{polino2018model}, the authors also proposed a random rounding scheme based on Equation (7). In this case, we sample $\xi \sim Bernoulli(xs-\floor{xs})$. This allows us to quantize the real values to multi-levels probabilistically. One important property of this random rounding function is that it is an unbiased estimator of the input, which means that $E(Q(x)) = x$. This reveals that this random rounding method equals to add noises into the training process.
\\

\noindent {\small \textbf{Challenges:}} Random rounding provides a way to inject noises into the training process. It can act as a regularizer and  enable conditional computation. However, with random rounding methods we need to estimate the gradient of the discrete neurons. Such estimation often has a high variance. This fact may cause oscillations in the loss function during training. The work of \cite{bengio2013estimating} provides an overview of possible solutions for estimating gradients for discrete neurons.

\begin{figure}[b]
	\centering	
	
	\includegraphics[height=1.6in]{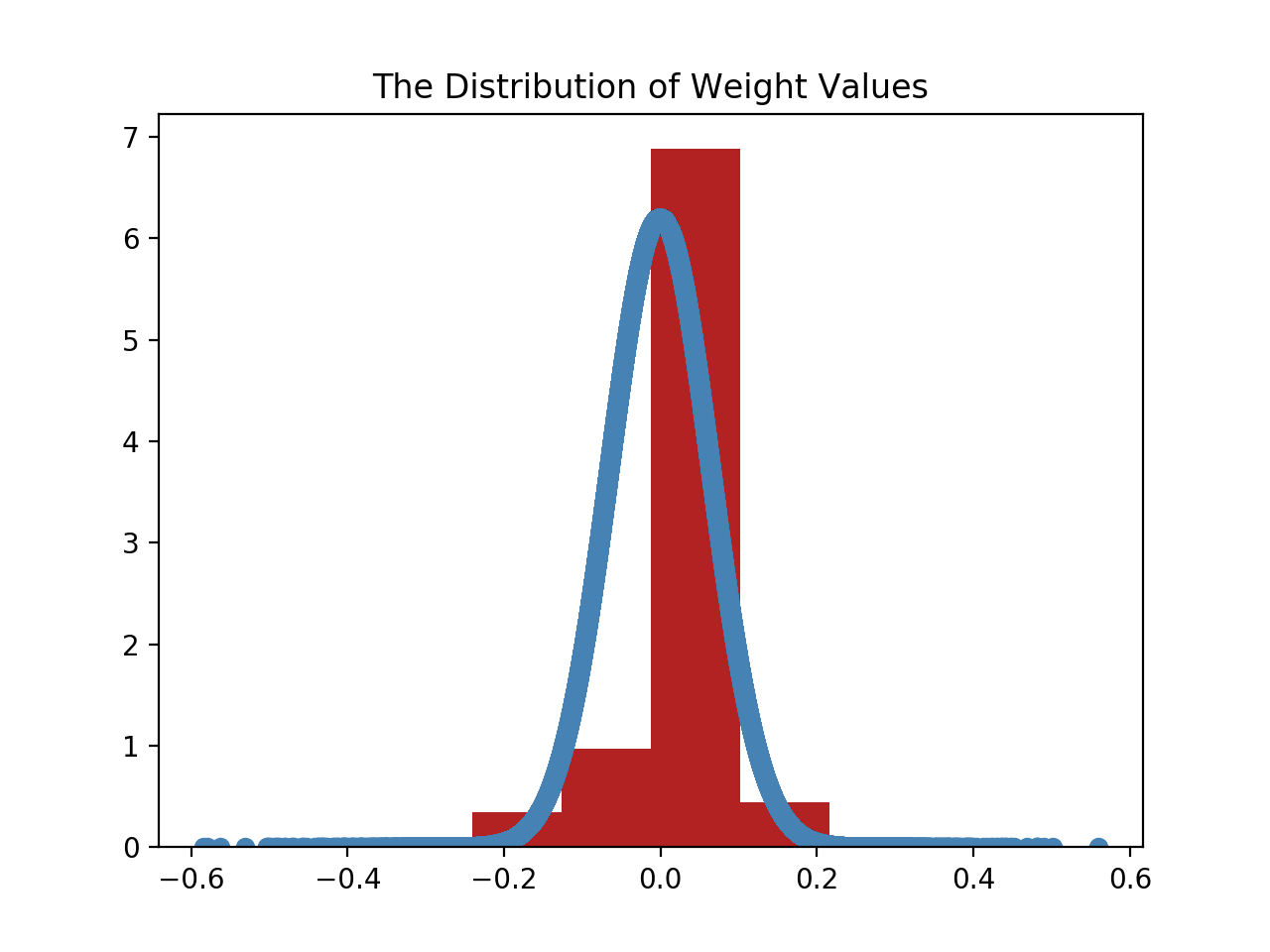}
	\caption{The histogram of weight values of LeNet-5 \protect\cite{lecun1998gradient} after training on MNIST. The blue line is the fitted Gaussian distribution. }
	\label{figure: dist}
\end{figure}

\setlength{\tabcolsep}{0.2em} 
{\renewcommand{\arraystretch}{1.2}
	\begin{table*}[!htb]
		\scriptsize
		\caption{Summarization of different quantization methods}
		\label{table:exp}
		\begin{center}
			\begin{tabular}{|c|c|c|c|}
				\hline
				Types & Techniques & Descrition & Characteristics \\ \hline
				
				& Rounding & \makecell{Using a quantization function to \\  convert continuous values into discrete values}  & \makecell{ Simple to implement, can achieve good performance, \\ often need to store the real values.} \\  
				
				Deterministic Quantization 	& Vector Quantization & Cluster the real values into subgroups & \makecell{Simple to implement, can explore structural redundancy, \\ only can be used to quantize pre-trained models.} \\ 	
				
				& Quantization as Optimization& \makecell{ Convert the quantization problem \\ into an optimization problem}  & \makecell{ Guaranteed to converge to a local minimum, \\ more difficult to implement } \\
				\hline 
				& Random rounding &  \makecell{Sampling quantized values according to given probabilities} & Simple to implement, introduce noises as regularizers  \\ 
				Stochastic Quantization	 &   &   &  \\
				& Probabilistic Quantization & 
				\makecell{Assume the weights are discretely distributed \\ or learn multi-modal posterior disribution over weights} & \makecell{Consider structural redundancy, automatic regularization, \\ easy to interpret  } \\
				
				\hline
			\end{tabular}
		\end{center}
	\end{table*}

\subsubsection{Probabilistic Quantization}
The weights in a trained network often follow some distributions. Figure \ref{figure: dist} shows the histogram of weight values in the LeNet \protect\cite{lecun1998gradient} after trained on MNIST dataset. It is obvious that most of the weight values are close to zero and the distribution is roughly Gaussian. The behavior of the weights inspired researchers to quantize the network from a probabilistic perspective.

In probabilistic quantization, the weights are assumed to be discretely distributed. A learning algorithm is used to infer the parameters of the distributions. In \cite{soudry2014expectation}, the authors developed the Expectation Back-propagation algorithm to train neural networks with binary or ternary weights. They first assumed some discrete prior distribution on the weights $p(\textbf{w}|D_0)$, and then updated the weights in an online setting based on the Bayesian formula,
\begin{equation}
	p(\textbf{w}|D_n) \propto p(y^n|x^n,\textbf{w}) p(\textbf{w}|D_{n-1})
\end{equation}

Above update rule is intractable in general, the authors adopted mean-field approximation and the Central Limit Theorem (CLT) to obtain an approximated solution.

In \cite{shayar2017learning}, the authors assumed that each weight is sampled independently from a multinomial
distribution and the loss function of the network is as follows,
\begin{equation}
	L(\mathbf{w}) = E_{w \sim \mathbf{w}}[\sum_{i=1}^N l(f(x_i, w), y_i)]
\end{equation}
This function is not differentiable due to the discreteness. The authors used local reparameterization trick \cite{kingma2015variational} and the Central Limit Theorem (CLT) to approximate the discrete distributions by a smooth Gaussian distribution. In this way, the gradients can be back-propagated through the discrete nodes.

Another type of probabilistic quantization is based on variational inference \cite{jordan1999introduction}. The main idea is to place a quantizing prior on the weights and then use variational inference to obtain the discrete posterior distribution of the weights. Assume a dataset $D = (x_n,y_n)^N_{n=1}$ and let $p(Y|X,\textbf{w})$ be a parameterized neural network model that predicts outputs $Y$ given inputs $X$ and parameters $\textbf{w}$. In Bayesian neural networks, we want to estimate the posterior distribution of the weights given the data: $P(\textbf{w}|D) = p(D|\textbf{w})p(\textbf{w})/p(D)$. $P(\textbf{w})$ is the prior distribution of the weights. To analytically solve the true posterior is intractable. One approach is to use variational inference algorithm to approximate the true posterior. The true posterior distribution is approximated by a parameterized distribution $q_{\phi}(\textbf{w})$. To find $q_{\phi}(\textbf{w})$, we need to minimize the Kullback-Leibler divergence between the true and the approximated posterior distribution: $D_{KL}(q_{\phi}(\textbf{w})||p(\textbf{w}|D)$. This optimization problem can be further converted to maximize the following ``evidence lower bound" (ELBO),
\begin{equation}
L(\phi) = \sum_{n=1}^N \mathbb{E}_{q_{\phi}(\textbf{w})}[\log p(y_n|x_n,\textbf{w})] - D_{KL}(q_{\phi}(\textbf{w}) || p(\textbf{w}))
\label{eq: elbo}
\end{equation}

The first term of the right-hand side of Equation (21) is the negative of reconstruction error, which means that maximize this term will ensure good predictive performance. The second term of the right-hand side of Equation (21) regularizes the approximated posterior to be close to the prior distribution. 

Traditional Bayesian neural networks do not involve quantization. In \cite{kingma2015variational}, the authors connected the variational training of Bayesian neural networks with dropout \cite{srivastava2014dropout}. In dropout training, Bernoulli noises or Gaussian noises are added to the weights. In traditional dropout training, the dropout rate $p$ is fixed. \cite{kingma2015variational} shown that adding multiplicative noise on weights is equivalent to learn adaptive dropout rate for each weight. If we add a Gaussian noise $\xi_{ij} \sim N(1, \alpha = \frac{p}{1-p})$ on each weight, then the weight is distributed as follows:
\begin{equation}
	w_{ij} = \theta_{ij}\xi_{ij} = \theta_{ij}(1+\sqrt{\alpha}\epsilon_{ij}) \sim N(\theta_{ij}, \alpha \theta_{ij}^2)
\end{equation}
In a Bayesian neural networks setting, the gradient of the weights can be computed as,
\begin{equation}
	\begin{split}
		\Delta_w \log p(y | x,w) & = \Delta_w \log p(y | x, \theta\hat{\xi})\\
		& = \Delta_w \int q(w|\theta, \alpha)]\log p(y | x,w)dw
	\end{split}
	\label{eq: dropout}
\end{equation}
where $\hat{\xi} \sim N(\xi|1, \alpha I)$. The gradient of ELBO with respect to the weights is Equation \ref{eq: dropout} plus the gradient of the KL divergence term. To exactly recover the ELBO loss, \cite{kingma2015variational} adopted a prior distribution on the weights which makes the KL divergence term in Equation \ref{eq: elbo}  does not depend on $\theta$ but on $\alpha$. This allows us to learn different dropout rates for different weights. \cite{molchanov2017variational} shown that we can prune the weights that have high dropout rates and still achieve good predictive performance.

In \cite{achterhold2018variational}, the authors introduces a "multi-spike-and-slab" prior which has multiple spikes at locations $c_k$, $k \in {1...K}$. After training, they found that most weights of low variance are distributed very closely around the quantization target values $c_k$ and can thus be replaced by the corresponding $c_k$ without significant loss in accuracy. Weights of large variance can be pruned. 
\\

\noindent {\textbf{Challenges:}} Probabilistic quantization can leverage the benefits of Bayesian neural networks which leads to very sparse models. However, it relies on a carefully chosen prior distribution of the weights and the model is often intractable. Meanwhile, some types of neural network models, such as recurrent neural networks, cannot be quantized under this framework. 

\subsection{Discussion}
The above quantization techniques enable us to quantize neural networks from different perspectives. We summarize all the techniques in Table \ref{table:exp}. The merits and drawbacks of these techniques can guide us to select the proper one in different situations. In general, deterministic quantization should be preferred if we want to quantize neural networks for hardware accelerations since we can specify the appropriate quantization levels in advance in order to run the quantized networks on dedicated hardware. This can give us predictive performance improvement on hardware. Round rounding enables us to quantize the weights in a data-dependent manner. This leads to conditional computation \cite{bengio2013estimating} that can increase the capacity of neural networks. Probabilistic quantization differs from deterministic quantization in that the quantized weights are more interpretable. We can understand the distributions of the weights with probabilistic quantization and gain more insights into how the network works. With probabilistic quantization, we can also have sparser models due to regularization effects of the Bayesian methods.

\section{Quantization of Network Components}

\subsection{Weight Quantization}
The motivation to quantize weights is clear: to reduce model size and accelerate training and inference process. Most of the methods we talked above can be used to quantize weights. In this section, we introduce more weight quantization strategies that we did not cover before.

\cite{anwar2015fixed} proposed a layer-wise quantization scheme to reduce the performance degradation. In \cite{kim2016bitwise}, the authors adopted a two-step pipeline. In the first step, the weights are compressed into the range of $[-1,1]$ and in the second step the compressed weights are used to initialize the parameters of a binary network. In \cite{zhou2017incremental}, the authors proposed incremental network quantization (INQ) which consists of three steps: weight partition, group-wise quantization and re-training. They quantized the weights in a group-wise manner to allow some groups of weights to compensate the accuracy loss due to the quantization of other groups. The work in \cite{gudovskiy2017shiftcnn} extended this method to power-of-two setting.

In \cite{lin2016fixed} the authors tried to find the optimal fixed point bit-width allocation across
layers. They examined how much noise  can be introduced by quantizing different layers. \cite{lin2017towards} approximated the full-precision weights with a linear combination of multiple binary bases. The results show that it is the first time that a binary neural network can achieve prediction accuracy comparable to its full-precision counterpart on ImageNet dataset. In \cite{moons2017minimum}, the authors studied how to develop energy efficient quantized neural network. The work in \cite{guo2017network} introduced network sketching to quantize a pre-trained model. The idea is to use binary basis to approximate pre-trained filters. They first proposed a heuristic algorithm to find the binary basis and then provided a refined version to better approximation. In \cite{amer2018bitregularized}, the authors proposed an end-to-end training framework to optimize original loss function, quantization error and the total number of bits simultaneously. However, the accuracy is not comparable to other quantized neural networks.
\\

\noindent \textbf{Challenges} 
\begin{itemize}
	\item Quantized weights make neural networks harder to converge. A smaller learning rate is needed to ensure the network to have good performance \cite{wu2018training}. 
	Determine how to control the stability of the training process in a quantized neural network with quantized weights is critical.
	
	\item Quantized weights make back-propagation infeasible since gradient cannot back-propagate through discrete neurons. Approximation methods are needed to estimate the the gradients of the loss function with respect to the input of the discrete neurons. Developing low-variance, unbiased gradient estimates is essential for the success of weight quantization.
	
	\item It is known that the weights in neural networks often follow some general structures. For an approach that trains quantized networks from scratch, how to quantize the weights locally while maintain their global structure is an issue. 
\end{itemize}

\subsection{Activation Quantization}
Quantized activations can replace inner-products with binary operations which can further speed up the network training. We can also reduce the much memory by avoiding full-precision activations. \cite{vanhoucke2011improving} quantized the activations to 8 bits. They used a sigmoid function which limits the activations to the range of [0, 1] and quantized the activations after training the network.
In \cite{courbariaux2015binaryconnect,rastegari2016xnor,zhou2016dorefa} the authors adopted a similar approach. They introduced a continuous approximation of the non-differentiable operator during back-propagation to enable the gradients can  back-propagate through the discrete neurons. More recently, \cite{cai2017deep} proposed an half-wave
Gaussian quantizer to approximate the ReLU unit. In the forward approximation, they used a half-wave Gaussian quantization function,
\begin{equation}
Q(x) = \begin{cases}
	q_i\quad \textnormal{if}\ x \in (t_i,t_{i+1}],\\
	0\quad x \le 0, \\
	\end{cases}	
\end{equation}
If use mean squared error to measure the performance, the optimal quantizers can be found as follows,
\begin{equation}
	Q^*(x) = \argmin_{Q}\mathbf{E}_x[(Q(x)-x)^2]
\end{equation}
They used batch normalization \cite{ioffe2015batch} and Lloyd's algorithm to find the optimal solution. During back-propagation, they further introduced three possible approximation method to avoid the gradient vanishing problem.

In \cite{mishra2017wrpn}, the authors proposed wide reduced-precision networks (WRPN) to quantize activation and weights. They found that activations actually occupy more memory than weights. They adopted a strategy that increases the number of filters in each layer to compensate the accuracy degradation due to quantization.
\\

\noindent \textbf{Challenges} 
\begin{itemize}
	\item There are some reasons that make the quantization of activations more difficult than that of weights \cite{cai2017deep}. The first one is that we need to back-propagate through the non-differentiable operators. Consider the back-propagation equation,
	\begin{equation}
	\frac{\partial L}{ \partial w_{ij}} = g'(a_j) \sum_{k} w_{jk} \frac{\partial L}{\partial a_k}x_i
	\end{equation}
	When we replace $g(a_j)$ with a binary operator, the derivative $g'(a_j)$ is almost zero everywhere which makes gradient descent algorithm infeasible. 
	
	\item 
	The quantized activations can lead to ``gradient mismatch" problem \cite{lin2016overcoming} which means that there is a discrepancy between the quantized activation with the computed backward gradient. 
	
\end{itemize}

\begin{figure} [!htbp]
	\subfigure[]{   
		\includegraphics[width=1.6in]{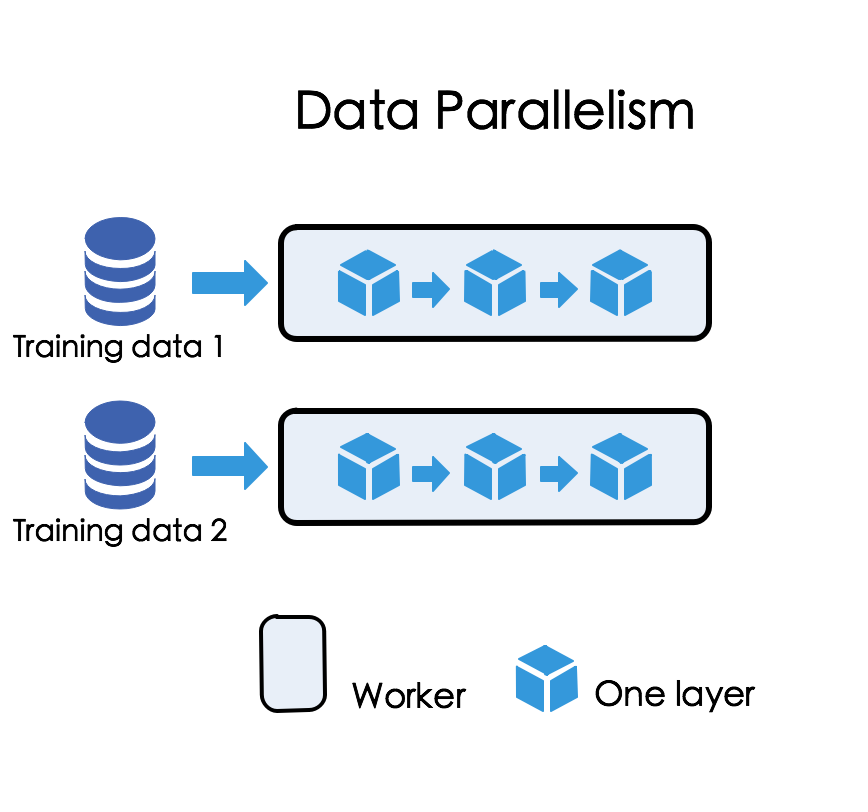}}
	\subfigure[]{   
		\includegraphics[width=1.6in]{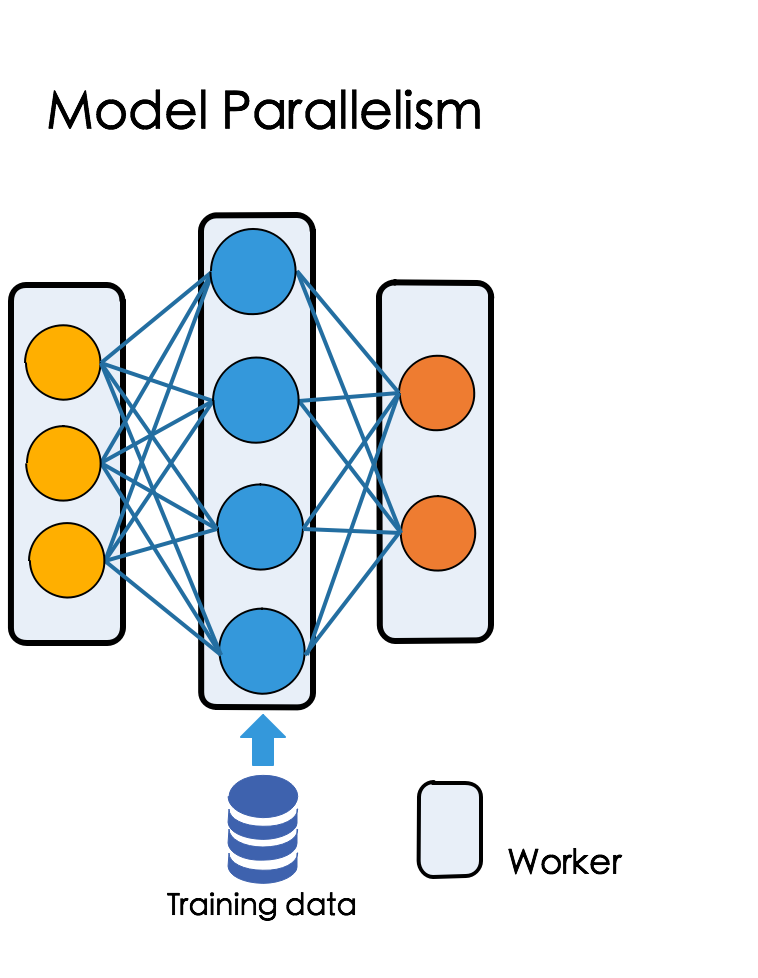}}
	
	\caption{Parallel training of deep neural networks. }  
	\label{fig: para}
\end{figure}

\setlength{\tabcolsep}{0.4em} 
{\renewcommand{\arraystretch}{1}
	\begin{table*}[!htb]
		\caption{Summarization of Quantization of Network Components}
		\label{table:exp}
		\begin{center}
			\begin{tabular}{|c|c|c|}
				\hline
				Components &   Benefits from Quantization  & Challenges \\ \hline
				
				&  Smaller model size &  Hard to converge with quantized weights  \\  
				
		Weights	&  Faster training and inference & Require approximate gradients   \\ 	
				&  Less energy & Accuracy degradation \\				
				\hline 
				
					&  Smaller memory footprint during training &   \\  
	Activations	&  Allows replacement of dot-products by bitwise operations &   ``Gradient mismatch" problem \\ 	
	&  Less energy&  \\		
	\hline

Gradients	&  Communication and memory savings in parallel network training  & Convergence requirement \\ 	
	
				\hline
				
			\end{tabular}
		\end{center}
	\label{table: components}
	\end{table*}

\subsection{Gradient Quantization}
Gradient quantization is a new branch of research in quantization of neural networks. The motivation to quantize gradients is to reduce the communication cost during distributed stochastic gradient descent (SGD) training of large neural networks.

In a distributed SGD training, one case is that each mini-batch data is spread over multiple computing nodes, this is called data-parallel training as shown in Fig \ref{fig: para} (a). Each node has a copy of the weights and need to compute a sub-gradient, and then broadcast its sub-gradient to all other nodes. Each node must accumulate the sub-gradients from other nodes to update the weights. This process causes a significant performance bottleneck because of the exchange of gradients. To solve this problem, \cite{seide20141} proposed to use 1-bit to represent the sub-gradient. This can reduce the bandwidth greatly. The authors reported a ten times speed-up compared with traditional approaches without a great loss in accuracy. In \cite{strom2015scalable} the authors proposed a threshold quantization method to quantize gradients. A fixed threshold is selected in advance. Gradients that are greater than the threshold are quantized to +1, and those less than the threshold are quantized to 0. \cite{alistarh2016qsgd} also considered the problem of gradient communication in the parallel SGD. They proposed Quantized SGD (QSGD) to allow each node to make a trade-off between the precision the gradients with the accuracy of the model. QSGD used the idea of random rounding to quantized the gradients to a set of discrete values and utilized lossless code to generate efficient encoding. In \cite{dryden2016communication}, the authors proposed a simple adaptive quantization method to select a proportion of gradients to be quantized and sent.

In a different setting, there is a centralized parameter server that performs gradient synchronization by accumulating all sub-gradients and averaging them to update the weights. This is called model-parallel training as shown in Fig \ref{fig: para} (b). The updated weights are sent back to each node to do computation. As the number of nodes increases, the communication cost becomes intolerable. \cite{wen2017terngrad} addressed this problem by introducing a method called TernGrad that quantizes the gradients into three levels $\{-1,0,1\}$. Before being sent to the centralized parameter server, each sub-gradient is quantized as follows,
\begin{equation}
\tilde{\Delta}_t = \textnormal{ternarize}(\Delta_t) = s_t \cdot \textnormal{Sign}(\Delta_t) \circ b_t
\end{equation}
where $s_t = \textnormal{max}(\textnormal{abs}(\Delta_t))$, $\circ$ is the Hadamard product and $b_t$ is a random binary vector that follows a Bernoulli distribution,
\begin{equation}
\begin{cases}
\hskip0.3emP(b_{tk} = 1| \Delta_t) = |\Delta_{tk}|/s_t, \\\
P(b_{tk} = 0| \Delta_t) = 1- |\Delta_{tk}|/s_t, \\
\end{cases}	
\end{equation}

\setlength{\tabcolsep}{0.4em} 
{\renewcommand{\arraystretch}{1}
	\begin{table*}[!htb]
		\caption{Summarization of fixed codebook and adaptive codebook quantization.}
		\label{table:exp}
		\begin{center}
			\begin{tabular}{|c|c|c|c|}
				\hline
				Approaches & Types &  Codebook & Representative works \\ \hline
				
				& Binarization & \{-1,1\} &  \cite{courbariaux2015binaryconnect} \\  
				
					& Scaled binarization &  \{-a, b\} &  \cite{rastegari2016xnor}  \\ 	
				Fixed codebook quantization	& Ternarization &  \{-1, 0, 1\}& \cite{hwang2014fixed} \\				
				& Scaled Ternarization &  \{-a,0,b\} &  \cite{zhu2016trained,kim2014x1000} \\
				
				& Powers of two & \{0, $\pm$1, $\pm$ $2^{-1}$,...,$\pm$ $2^{-L}$ \} & \cite{tang1993multilayer} \\ 
				\hline 
				Adaptive codebook quantization	 & Soft Quantization &  Learned from data & \cite{achterhold2018variational}  \\
				&  Hard Quantization & Learned from data & \cite{gong2014compressing,choi2016towards}
				\\
				\hline
			\end{tabular}
		\end{center}
	\end{table*}
In this way, th communication cost between the server and workers can be reduced by about 20 $\times$ compared with sending full-precision gradients. 

In a single-machine environment, we can also gain benefits by quantizing gradients. In order to reduce the computational cost in the backward pass, the work in \cite{rastegari2016xnor} quantized the gradients into 2-bits to enable an efficient training process. In \cite{zhou2016dorefa}, the authors also quantized the gradients during back-propagation. They found that using a random rounding method is very important to make quantized gradients work well. They designed the following $k$-bit quantization function,
\begin{equation}
\tilde{f}_{\gamma}^{k}(dr) = 2\textnormal{max}_0(|dr|)[\textnormal{quantize}_k(\frac{dr}{2\textnormal{max}_0(|dr|)+\frac{1}{2}}) -\frac{1}{2}]	
\end{equation}
where $dr = \frac{\partial c}{\partial r}$ is the gradient of the output $r$ in some layer and $\textnormal{quantize}_k$ is used to quantize a real number input $r_i \in [0,1]$ into a $k$-bit output number $r_o \in [0,1]$,
\begin{equation}
r_o = \frac{1}{2^k-1}\textnormal{round}((2^k-1)r_i)
\end{equation}
They also added additional noises during the training process to compensate for the loss of accuracy due to quantization.
\\

\noindent \textbf{Challenges} 
\begin{itemize}
	\item The magnitude and sign of gradients are both important for updating the weights. To quantize gradients, we must address the question of how to take both factors into account.
	
	\item A naive way to quantize gradients may not work well in practice since it may violate the conditions needed for stochastic gradient descent algorithm to converge. More sophisticated methods are needed in this case.
\end{itemize}

\subsection{Discussion}
We summarize the benefits and challenges of quantizing different network components in Table \ref{table: components}. Achieving highly efficient quantized neural networks calls for a systematic solution to quantize weights, activations and gradients. Quantized weights and activations occupy less memory compared with full-precision counterparts. Meanwhile, the training and inference speed can be greatly accelerated since the dot-products between weights and activations can be replaced by bitwise operations. Quantized gradients can reduce the overhead of gradient synchronization in parallel neural network training. In a single worker scenario, quantized gradients can accelerate back-propagation training as well as requiring less memory.

\section{A Comparison of Two Quantization Methodologies}

\subsection{Fixed codebook quantization}
In fixed codebook quantization, the codebook is predefined. For example, \cite{courbariaux2015binaryconnect} quantized the weights of the network to $\{-1,1\}$. \cite{hwang2014fixed} assumed the codebook is $\{-1,0,1\}$. In \cite{rastegari2016xnor}, the authors further relaxed the constraints to quantize the weights to $\{a, -a\}$. In a more general setting, the weights are quantized into power-of-two numbers that can make the digital implementation of neural networks much faster \cite{tang1993multilayer,gudovskiy2017shiftcnn}. 

In order to achieve fixed codebook quantization, we must define a codebook first. How the codebook is designed has a dramatic impact on the performance of the quantized network. A small codebook means that we can only search the parameters in a limited space, which makes the optimization problem very hard. With predefined codebook, it is sometimes necessary to modify the backward step to enable the gradients flow through the discrete neurons. This causes the ``gradient mismatch" problem as we discussed in Section 4.2. Approximation is often needed in this case. 

\subsection{ Adaptive Codebook Quantization}
In adaptive codebook quantization, the codebook is learned from the data. Vector quantization and probabilistic quantization are two possible methods to achieve adaptive codebook quantization. In vector quantization, in order to learn the codebook, we must let the real values minimize some sort of distortion measure and cluster them into different buckets. In probabilistic quantization, the codebook can be inferred from the posterior distributions of the weights.

Two kinds of adaptive codebook quantization exist: hard quantization and soft quantization. In hard quantization, the real value is assigned exactly to be one of the discrete values. In soft quantization the real value is assigned to be some discrete value according to a probability distribution. The soft quantization is mostly inspired by the idea of weight sharing \cite{nowlan1992simplifying} in which the distribution of weight values is modeled as a mixture of Gaussians. Adaptive quantization is more flexible than fixed codebook quantization but the final codebook may need more bits to represent. The benefit of adaptive quantization is that it can avoid ad hoc modifications to the training algorithm.
\\

\section{Quantized Neural Networks: Case Studies}
Neural networks can be quantized during and after training. In this section we introduce some recently proposed quantized neural networks that utilize these this perspective.

\subsection{Quantization During Training}

\noindent \textbf{BinaryConnect} \cite{courbariaux2015binaryconnect} BinaryConnect is a method that leverages binary weights during the forward and backward pass. As far as we known, it is the first time that a binary network can achieve near state-of-art results on datasets such as MNIST and CIFAR-10. BinaryConnect uses the Sign(x) function to binarize the weights during the forward pass. The real-valued weights are also kept to do parameter update in the backward pass. The algorithm is shown in Algorithm \ref{Algorithm: bc}. 
\\

\begin{algorithm}[!h]
	\textbf{Notation:} $L$ is the number of layers in the network. $\textbf{w}_{t-1}$ is the weight at time $t-1$. $b_{t-1}$ is the bias at time $t-1$. $a_k$ is the activation of layer $k$. $E$ is the loss function.
	
	\textbf{Input:} a mini-batch of data (inputs, labels),  a learning rate $\eta$.

	\textbf{Forward pass:}
	\begin{itemize}
		\item $\textbf{w}_b$ $\leftarrow$ \textnormal{binarize}($\textbf{w}_{t-1}$)
		\item For $k=1$ to $L$, compute $a_k$ based on $a_{k-1}$, $\textbf{w}_b$ and $b_{t-1}$. 
	\end{itemize}
	
	\textbf{Backward pass:}
	
	\begin{itemize}
		\item Compute the gradient of the output layer $\frac{\partial E}{\partial a_L}$
		\item For $k=L$ to 2, compute $\frac{\partial E}{\partial a_{k-1}}$ based on $\frac{\partial E}{\partial a_k}$ and $\textbf{w}_b$. 
	\end{itemize}
	
	\textbf{Parameter update:}
		\begin{itemize}
			\item Compute $\frac{\partial E}{\partial \textbf{w}_b}$ and $\frac{\partial E}{\partial b_{t-1}}$ based on $\frac{\partial E}{\partial a_k}$ and $a_{k-1}$
			\item $\textbf{w}_t$ $\leftarrow$ \textnormal{clip}($\textbf{w}_{t-1}$ $-$ $\eta$ $\frac{\partial E}{\partial \textbf{w}_b}$)
			\item $b_t$ $\leftarrow$ $b_{t-1}$ $-$ $\eta$ $\frac{\partial E}{\partial b_{t-1}}$
		\end{itemize}
	
	\caption{The training process of BinaryConnect}
	\label{Algorithm: bc}
\end{algorithm}

\noindent \textbf{XNOR-Net} \cite{rastegari2016xnor} XNOR-Net is the first attempt to present an evaluation
of binary neural networks on large-scale datasets like ImageNet. They use a different binarization method compared with BinaryConnect. XNOR-Net binarizes inputs, weights, activations and gradients together which can greatly accelerate the network training and inference process. XNOR-net binarizes the gradients in the backward pass with a slightly drop in accuracy. The algorithm is shown in Algorithm \ref{Algorithm: xnor}.
\\

\begin{algorithm}[!t]
	\textbf{Notation:} $L$ is the number of layers in the network. $\textbf{w}_{t-1}$ is the weight at time $t-1$. $b_{t-1}$ is the bias at time $t-1$. $a_k$ is the activation of layer $k$. $n$ is the number of elements in a filter. $E$ is the loss function. 
	
	\textbf{Input:} a minibatch of data (inputs, labels),a learning rate $\eta_t$.
	
	\textbf{Forward pass:}
	\begin{enumerate}
		\item for $l=1$ to $L$ do:
		\item \quad for $k^{th}$ filter in $l^{th}$ layer do:
		\item  \qquad	$\textbf{a}^{lk}$ = $\frac{1}{n} \lVert \textbf{w}^{lk}_{t-1}  \rVert_{l_1}$ 
		\item  \qquad	$\textbf{b}^{lk}$ = Sign($\textbf{w}^{lk}_{t-1}$)
		\item \qquad $\textbf{w}_b^{lk}$ = $\textbf{a}^{lk}\textbf{b}^{lk}$
		\item Compute activations based on binarized filters
	\end{enumerate}

	\textbf{Backward pass:}
	\begin{enumerate}
		\item Compute backward gradient $\frac{\partial E}{\textbf{w}_b}$ based on $\textbf{w}_b$
	\end{enumerate}

	\textbf{Parameter update:}
	\begin{enumerate}
		\item Update $\textbf{w}_{t}$ using $\textbf{w}_{t-1}$ and $\frac{\partial E}{\textbf{w}_b}$
	\end{enumerate}

	\caption{The training process of XNOR-net}
	\label{Algorithm: xnor}
\end{algorithm}

\noindent \textbf{DoReFa-Net} \cite{zhou2016dorefa} DoReFa-Net further improves XNOR-net by using more sophisticated rounding mechanism. They claim that it is the first time that during the backward pass quantized gradients with less than 8 bits can work successfully. DoReFa-Net can binarize weights, activations and gradients to arbitrary bit-width. The forward pass and backward pass can both be greatly accelerated. We omit the details of the algorithm here due to space limitation.
\\

\noindent \textbf{ABC-Net} \cite{tang2017train} In ABC-Net, the authors studied carefully why the previous binary networks may fail. And they proposed following strategies to alleviate the potential problems,
\begin{itemize}
	\item Use a smaller learning rate to prevent the frequent changes of the directions of weight values.
	\item Use PReLU \cite{he2015delving} rather than ReLU as the activation function.
	\item Use the following regularization function rather than a $L_2$ regularization,
	\begin{equation}
		L = \lambda \sum_{l=1}^L \sum_{i=1}^{N_l} \sum_{j=1}^{M_l} (1-(\textbf{w}_{l,ij})^2)		
	\end{equation}
	where $L$ is the number of layers. $N_l$ and $M_l$ are the dimensions of the weight matrix in layer $l$.
 \end{itemize}

In order to successfully quantize the final layer, the authors also added an additional scale layer after the binarized final layer to improve the compression rate of the algorithm.  \\

Figure \ref{fig: binary} shows the general procedures of quantizing neural networks during training. It can be noted that the full-precision weights must be saved in the training phase which can cause a large memory overhead.

\begin{figure}[h]
	\begin{center}
	\includegraphics[width=3.5in]{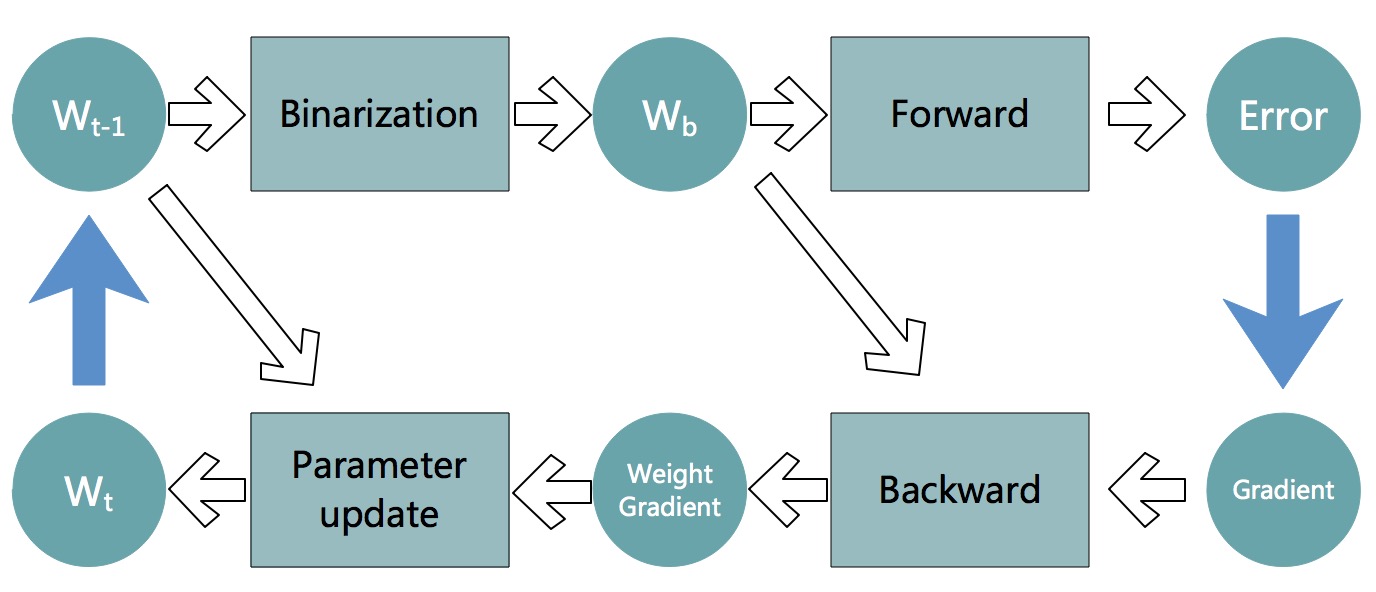} 
	\end{center}
	\caption{the general procedures of quantizing neural networks during training. $W_{t-1}$ and $W_{t}$ are real-valued weights while $W_b$ is the quantized weights.}
	\label{fig: binary}
\end{figure}

\setlength{\tabcolsep}{0.6em} 
{\renewcommand{\arraystretch}{1.3}
	
	\begin{table*}[!h]
		\caption{ Top 1 error rates of different quantized neural networks on the validation set.}
		\label{table:exp}
		\begin{center}
			\begin{tabular}{|c| c|c|c|c|c|c|}
				\hline
				Datasets	& \backslashbox{Methods}{Networks}	& 	MLP & AlexNet & VGGNet-16 & GoogleNet & ResNet-x \\ \hline \hline
				&	Binary-Connect \cite{courbariaux2015binaryconnect} &  1.20\% &  ---  & --- & --- & --- \\ 
				
				&	BNN \cite{hubara2016binarized}  & 0.70\%  &  ---  &  --- &--- & ---\\
				
				MNIST	& LAB \cite{hou2016lossaware}  & 1.18\%  &  ---  &  --- &--- & ---\\  
				& Gated XNOR Networks \cite{deng2017}  & 0.68\%  &  ---  &  --- &--- & ---\\

				& LR-net \cite{shayar2017learning}  & 0.50\%  &  ---  &  --- &--- & ---\\ 
				& WAGE \cite{wu2018training}  & 0.40\%  & ---  &  --- &--- &  ---\\		
				
				\hline
				\hline
				
				&	Binary-Connect \cite{courbariaux2015binaryconnect}&  --- & 8.4\% & --- &--- & --- \\ 
				
				&	BNN \cite{hubara2016binarized} &  ---  &  10.2\% &--- & --- &--- \\ 
				
				& TTQ \cite{zhu2016trained} & --- & ---    &--- &  ---& 6.4\% (x = 56) \\ 
				
				Cifar-10	&	XNOR-Net \cite{rastegari2016xnor}  &   ---  & 10.2\% & ---& ---& ---\\

				& Gated XNOR Networks \cite{deng2017}  & ---  &  7.5\%  &  --- &--- & ---\\ 
				
				& LR-net \cite{shayar2017learning}  & ---  &  ---  &  6.74\% &--- & ---\\ \label{key}
				&	LAT \cite{hou2018lossaware}  &   ---  & 10.38\% & ---& ---& ---\\ 
				
				& WAGE \cite{wu2018training}  & ---  & ---  & 6.78\% &--- &  ---\\

				\hline
				\hline
				
				&	Binary-Connect \cite{courbariaux2015binaryconnect}  &  --- & 64.6\%  &  ---& --- & ---\\ 
				
				&	DeepCompression \cite{han2015deep} &  ---  & 42.8\%  & 31.17\% & --- & ---\\
				
				&	BNN \cite{hubara2016binarized}  &  --- &  72.1\%  & --- & --- & ---\\ 
				
				&	TTQ \cite{zhu2016trained} &   ---  & 42.5\%  & --- & --- & 33.4\% (x = 18) \\ 
				
				&	DOREFA-Net \cite{zhou2016dorefa}  &  ---   &  44.4\% &---  &--- & 40.8\% (x=18) \\ 
				
				ImageNet	&	XNOR-Net \cite{rastegari2016xnor} &   ---  &  55.8\%  &---   &  34.5\%  & 48.8\% (x = 18)\\ 
				
				&	INQ \cite{zhou2017incremental}&  ---  & 42.6\%  & 29.2\% & 31.0\% & 7.6\% (x=50) \\
				
				& BinaryNet \cite{tang2017train} & ---  &  53.4\%  & --- & ---&--- \\ 
				
				& ABC-Net \cite{lin2017towards} & ---  &  ---  & --- &--- &  35.5\% (x=18) \\ 
				
				& WRPN (binary) \cite{mishra2017wrpn} & ---  & 51.7\%  & --- & 34.98\%  &  30.15\% (x=34)  \\   
	
				& LR-net \cite{shayar2017learning}  & ---  &  ---  &  --- &--- &  36.5\% (x=18)\\ 
				
				& WAGE \cite{wu2018training}  & ---  & 51.6\%  &  --- &--- &  ---\\					

				\hline
			\end{tabular}
		\label{results}
		\end{center}
	\end{table*}

\subsection{Quantization After Training}

\noindent \textbf{DeepCompression} \cite{han2015deep} DeepCompression is a three stage pipeline that can reduce the memory requirement of network by 35$\times$ to $49\times$ without accuracy degradation. The algorithm first prunes the unimportant connections. Then the remaining weights are quantized to discrete values. Finally, Huffman coding is used to encode the weight values. Figure \ref{figure: dp} shows the pipeline of DeepCompression,

\begin{figure}[h]
	\centering	
	
	\includegraphics[width=3.5in]{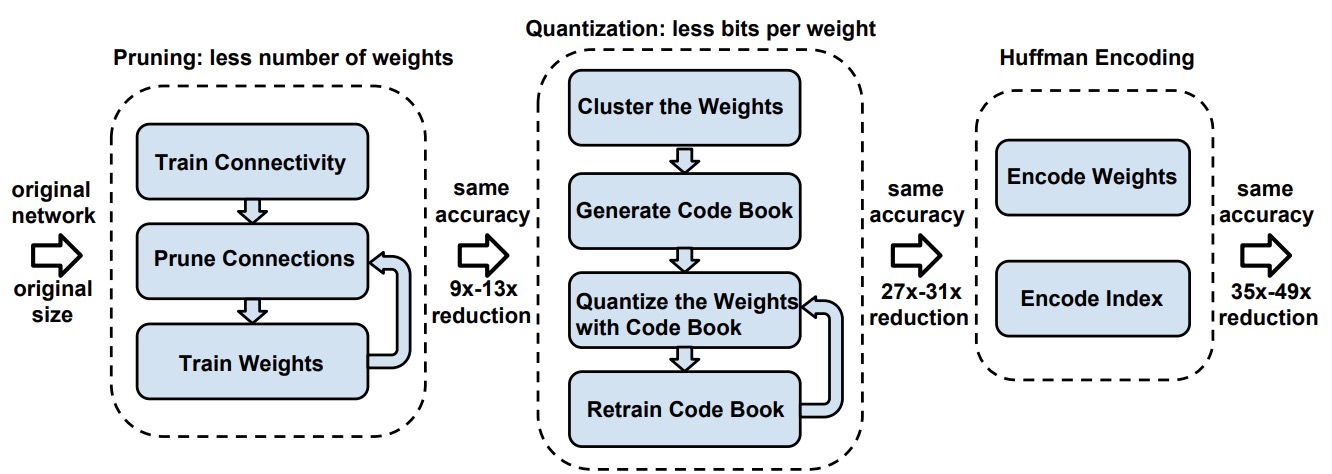} 
	
	\caption{The pipelines of DeepCompression \protect\cite{han2015deep}.}
	\label{figure: dp}
\end{figure}

In order to compensate the loss of accuracy, DeepCompression also retrains the remaining connections and the quantized centroids. In this way, high compression rate can be achieved while maintaining a good network performance.   \\

\noindent \textbf{Entropy-constrained scalar quantization (ECSQ)} \cite{choi2016towards}  Entropy-constrained scalar quantization (ECSQ) was proposed to improve the performance the vector quantization in 2016. In this work, the authors use the second-order information of the loss function to measure the importance of different weights. The loss function is expanded via Taylor series as follows,
\begin{equation}
	\delta E(\textbf{w}) \approx \frac{\partial E( \textbf{w})}{\partial \textbf{w}}^T \delta \textbf{w} + \frac{1}{2}\delta \textbf{w}^TH(\textbf{w})\delta \textbf{w}
	\label{hessian}
\end{equation}
where $H(\textbf{w})$ is the Hessian matrix. To connect Equation (\ref{hessian}) with network quantization, the authors approximated the Hessian matrix as a diagonal matrix and treated $\delta \textbf{w}$ as the quantization error. The loss due to quantization can be expressed as,
\begin{equation}
	\delta E(\textbf{w}) \approx \frac{1}{2} \sum_{i=1}^N h_{ii}(\textbf{w})|\textbf{w}_i - \bar{\textbf{w}}_i|^2
\end{equation} 
where $\tilde{\textbf{w}}$ is the quantized version of real-valued weights $\textbf{w}$. In the k-means clustering step, the weights are weighted by the corresponding entries in the Hessian matrix. The results show that this method can achieve nearly the same accuracy level as a full-precision network on ImageNet dataset.
\\

\noindent \textbf{Incremental network quantization (INQ)} \cite{zhou2017incremental} The work in \cite{zhou2017incremental} proposed Incremental network quantization (INQ). INQ consists of three independent operations: weight partition, group-wise quantization and re-training. In the step of weight partition, the weights in each layer are divided into two groups. One group of weights are quantized while the another group of weights are kept with full-precision values. The network is retrained with the remaining full-precision weights to compensate for the loss due to quantization. This process is continued until all the weights are quantized. Compared with other quantization methods, this approach combines the benefits of quantizing during training and quantizing after training.

\subsection{Performance Comparison of Different Quantized Neural Networks} 
We report the performance of different quantized neural networks in Table \ref{results}. All the results are directly taken from the original papers. The performance of quantized neural networks improved rapidly in the recent years and now can achieve near the state-of-the-art results on ImageNet with binarized weights. We have following observations,

\begin{itemize}
	\item We can achieve much higher accuracy with quantized neural networks if we use more bits to represent weights. In \cite{shayar2017learning}, the authors found that binary networks are much harder to train compared with a ternary counterpart.
	
	\item In general, the methods that quantize networks after training obtain better results as compared with those that quantize during training. This is understandable since in the case of quantizing after training we have well pre-trained models as reference.
	
	\item For some datasets and architectures, there is still a performance gap between the quantized neural networks and full-precision ones. 
\end{itemize}

\section{Why Does Quantization Work?}
Deep neural networks have a huge number of parameters, but not all parameters are of equal importance. As pointed out in \cite{denil2013predicting}, in the best cases more than 95\% of the parameters in a neural network can be predicted without a drop in predictive performance. This means that we can use simpler parameterization to maintain the expressive power of deep neural networks. Recent work in model compression \cite{molchanov2017variational} suggests that nearly 99\% of weights can be pruned in some types of neural networks.

Deep neural networks are also robust to noise \cite{sung2015resiliency,merolla2016deep}. Adding noise to weights or inputs sometimes can achieve better performance \cite{srivastava2014dropout}. Random noise acts as regularizers which can potentially generalize the network better. In a quantized neural network, low-precision operations can be regarded as noise which may not hurt the network performance. Recent theories  \cite{NIPS2017_7163,DBLP:journals/corr/AndersonB17} suggest that quantized neural networks still maintain many important properties of full-precision ones which guarantees their performance.

Despite the success of many quantized neural networks on real datasets, the theoretical understanding is still very limited. \cite{NIPS2017_7163} analyzes the convergence properties of stochastic gradient descent (SGD) when the weights are quantized. The authors analyzed the convergence property of BinaryConnect \cite{courbariaux2015binaryconnect}. They found that if we assume the loss function is \textit{L}-\textit{Lipschitz} smooth, the loss of the  BinaryConnect network will converge at a rate linear in $\Delta$ to the loss of a full-precision network in expectation, where $\Delta$ is the resolution of the quantization function. The authors further gave some results on non-convex cases. 

\cite{DBLP:journals/corr/AndersonB17} analyzed the properties of binarized neural network from a geometrical perspective. They found that the binarization operation preserves some important properties of the full-precision networks, i.e,
\begin{itemize}
	\item Angle Preservation Property: They found that binarization almost preserves the direction of full-precision high-dimensional vectors. The angle between a random normal vector with its binarized version converges to 37$\degree$.
	
	\item Dot Product Proportionality Property: They showed that the dot products of the activations with the pre-binarization and post-binarization weights are approximately proportional to each other, i.e., $a \cdot w^b \sim a \cdot w^c$. Where $a$ is the activation of one layer, $w^b$ is the binarized version of full-precision weight $w^c$. $a \sim b$ means that $a = cb$, where $c$ is a scalar. 
\end{itemize} 

The implication is that although binarization may change the numerical values dramatically, the statistical properties of the forward computation are nearly kept. 

\begin{figure}[h]
	\centering	
	
	\includegraphics[height=1.3in]{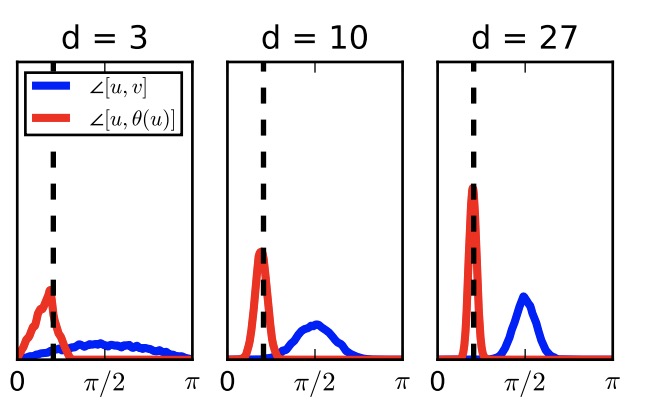} 
	
	\caption{ The red curves are distribution of angles a random vector of dimension $d$ and its binarized version. The blue curves are are distribution of angles two random vectors  \protect\cite{DBLP:journals/corr/AndersonB17}.}
	\label{figure: angle}
\end{figure}

\section{Future of Quantized Neural Networks}
Quantized neural networks make it practical to deploy deep neural network models into production stack. This enables embedded system based deep learning applications. However, there is still a large gap between the performance of quantized neural networks and full-precision neural networks. To bridge this gap, more sophisticated methods must be developed. Nearly all the works about quantized neural networks focus on feed-forward networks or convolutional neural networks and classification task. Recently, some researchers have looked at recurrent neural networks \cite{ott2016recurrent,hou2016lossaware,he2016effective,clark2017contextual,hou2018lossaware,xu2018alternating} and other tasks such as semantic segmentation \cite{wen2016training}, video processing \cite{o2016sigma} and so on. We believe that the wide use of deep neural networks will drive researchers to develop more task-specific quantized neural networks.

We consider the following possible directions for the next steps:
\begin{itemize}
	\item Develop more sophisticated rounding mechanism to train quantized neural network from scratch. One possible approach is to use the structure information of the weights to guide the rounding process.
	
	\item Design quantized neural networks for tasks such as natural language processing, speech recognition and so on. Due to the varieties of deep learning models, a generally applicable quantization method is necessary. \\
	
	\item Develop theoretical guidance for quantizing neural networks. 
\end{itemize}

\section{Conclusion}
In this paper, we have provided a comprehensive survey on the recent progress of quantized neural networks. We have traced back to the origins of the research of quantized neural networks and presented many newly developed methods. Both theories and applications of these methods are surveyed. We have pointed out some potential challenges in quantizing neural networks and have gave some general advice. We also identified several potential research directions. Quantized neural networks promote the application of deep learning models in mobile devices and embedded systems. We expect that they will make a significant impact in the future.

\bibliographystyle{named}
\bibliography{ijcai17}

\end{document}